\documentclass[runningheads]{llncs}

\usepackage[T1]{fontenc}
\usepackage{graphicx}
\usepackage{amsmath}
\usepackage{amssymb}
\usepackage{booktabs}
\usepackage{multirow}

\begin{document}

\title{Decoding Defensive Coverage Responsibilities in American Football Using Factorized Attention Based Transformer Models}
\titlerunning{Decoding Football Defense with Factorized Attention Transformers}

\author{Kevin Song\inst{1} \and
Evan Diewald\inst{1} \and
Ornob Siddiquee\inst{1} \and
Chris Boomhower\inst{1} \and
Keegan Abdoo\inst{2} \and
Mike Band\inst{2} \and
Amy Lee\inst{2}}

\authorrunning{Song et al.}

\institute{Amazon Web Services, Seattle, WA, USA \and
National Football League, New York, NY, USA}

\maketitle

\begin{abstract}
Defensive coverage schemes in the National Football League (NFL) represent complex tactical patterns requiring coordinated assignments among defenders who must react dynamically to the offense's passing concept. This paper presents a factorized attention-based transformer model applied to NFL multi-agent play tracking data to predict individual coverage assignments, receiver-defender matchups, and the targeted defender on every pass play. Unlike previous approaches that focus on post-hoc coverage classification at the team level, our model enables predictive modeling of individual player assignments and matchup dynamics throughout the play. The factorized attention mechanism separates temporal and agent dimensions, allowing independent modeling of player movement patterns and inter-player relationships. Trained on randomly truncated trajectories, the model generates frame-by-frame predictions that capture how defensive responsibilities evolve from pre-snap through pass arrival. Our models achieve approximately 89\%+ accuracy for all tasks, with true accuracy potentially higher given annotation ambiguity in the ground truth labels. These outputs also enable novel derivative metrics, including disguise rate and double coverage rate, which enable enhanced storytelling in TV broadcasts as well as provide actionable insights for team strategy development and player evaluation.

\end{abstract}

\section{Introduction}

Defensive coverage schemes in the National Football League (NFL) represent some of the most complex tactical patterns in modern team sports, involving coordinated assignments among as many as eight defensive players who must react dynamically to offensive formations and play development while maintaining specific coverage responsibilities~\cite{yurko_nfl_2025,dutta_unsupervised_2020,song_explainable_2023}. The ability to automatically predict individual coverage assignments, identify target defenders, and forecast receiver-defender matchups has profound implications for tactical analysis, player evaluation, and strategic decision-making. However, modeling defensive strategies is challenging, and previously, it was found that defensive backs are the most challenging position to model trajectories, noting their complex trajectory patterns involving long-distance travel and frequent direction changes based on ball perception and receiver movement~\cite{cheong_prediction_2021,lee_predicting_2016}.

Despite significant advances in sports analytics, current approaches to defensive coverage analysis face fundamental limitations. Previously, a CNN-LSTM model was introduced to predict coverage classification system at the play level, reaching 88.9\% accuracy across eight coverage types. However, this system focuses primarily on post-hoc coverage classification rather than predictive modeling of individual player assignments and matchup dynamics. Recent advances in multi-agent trajectory modeling have demonstrated the potential for sophisticated defensive analysis. The scene transformer architecture introduced unified approaches for multi-agent trajectory prediction using factorized attention mechanisms while AutoBots achieved state-of-the-art performance on multi-agent prediction tasks through interleaved temporal and social attention mechanisms~\cite{ngiam_scene_2022,girgis_latent_2022}. These architectures demonstrate strength in handling variable numbers of agents and complex interaction patterns, making them well-suited for NFL applications where player formations and tactical schemes vary dynamically. In addition, in other sports such as basketball, factorized attention architecture was applied to basketball plays to identify offensive plays~\cite{wang_hooptransformer_2024}.

Here, in this paper, we apply factorized attention transformer models on NFL multi-agent play tracking data to predict individual coverage assignments, receiver-defender matchups, and target defenders, and demonstrate the effectiveness and applicability of these models. We also demonstrate how these models together can provide additional derivative metrics, such as disguise rate and double coverage, that elucidate the effectiveness of defensive backs in American football.

\section{Related Work}

\subsection{Football Coverage Scheme Classification}

This effort builds on earlier collaborative work between the NFL Next Gen Stats (NGS) team and the Amazon ML Solutions Lab to classify NFL defensive schemes at the team coverage level (e.g. Cover 1, Cover 3)~\cite{song_explainable_2023}. Their ensemble CNN-attention model achieved 88.9\% accuracy across eight coverage types using frame-level tracking data, with temporal dynamics via self-attention being critical for capturing complex player interactions as plays unfold.

\subsection{Architectures for Modeling Spatio-Temporal Sequences}

Spatio-temporal predictive models are widely used across robotics~\cite{lee_predicting_2016}, traffic flow~\cite{deng_transposed_2022}, video analytics~\cite{noauthor_predrnn_nodate}, and sports~\cite{capellera_footbots_2024} to model causal relationships between multiple agents over time. The HoopTransformer architecture~\cite{wang_hooptransformer_2024} demonstrated effective factorized attention for NBA play classification by applying attention along the agent axis at each timestep, then along the time axis for each agent independently. This factorized approach efficiently aggregates spatio-temporal relationships in multi-agent scenarios, making it well-suited for NFL applications.

\section{Methodology}

\subsection{Model Architecture}

Our model architecture implements a factorized attention mechanism that is well suited for multi-agent trajectory data. Unlike traditional transformer attention mechanisms that apply full spatio-temporal attention, this architecture separates attention across temporal and agent dimensions for improved understanding of multi-agent relationships in both dimensions~\cite{wang_hooptransformer_2024}.

In addition to the attention-based encoding of player trajectories, our models also incorporate categorical metadata like player positions and a binary indicator of team (offense or defense) via learnable embeddings. These static features are fused with the outputs of the transformer to produce latent vectors corresponding to each player, which are then fed into a final classification head for the appropriate task. For coverage assignment specifically, additional, situational features are added, such as down, distance, yard line, time left in the game, and importantly, the overall team coverage scheme. During training, the ground truth team coverage scheme label is provided directly as an input feature. At inference time, a separate model~\cite{song_explainable_2023} predicts the team coverage scheme, and this prediction is used as the input feature in place of the ground truth label. These features are then broadcast into the latent player representations described above.

\begin{figure}[t]
\centering
\includegraphics[width=\textwidth]{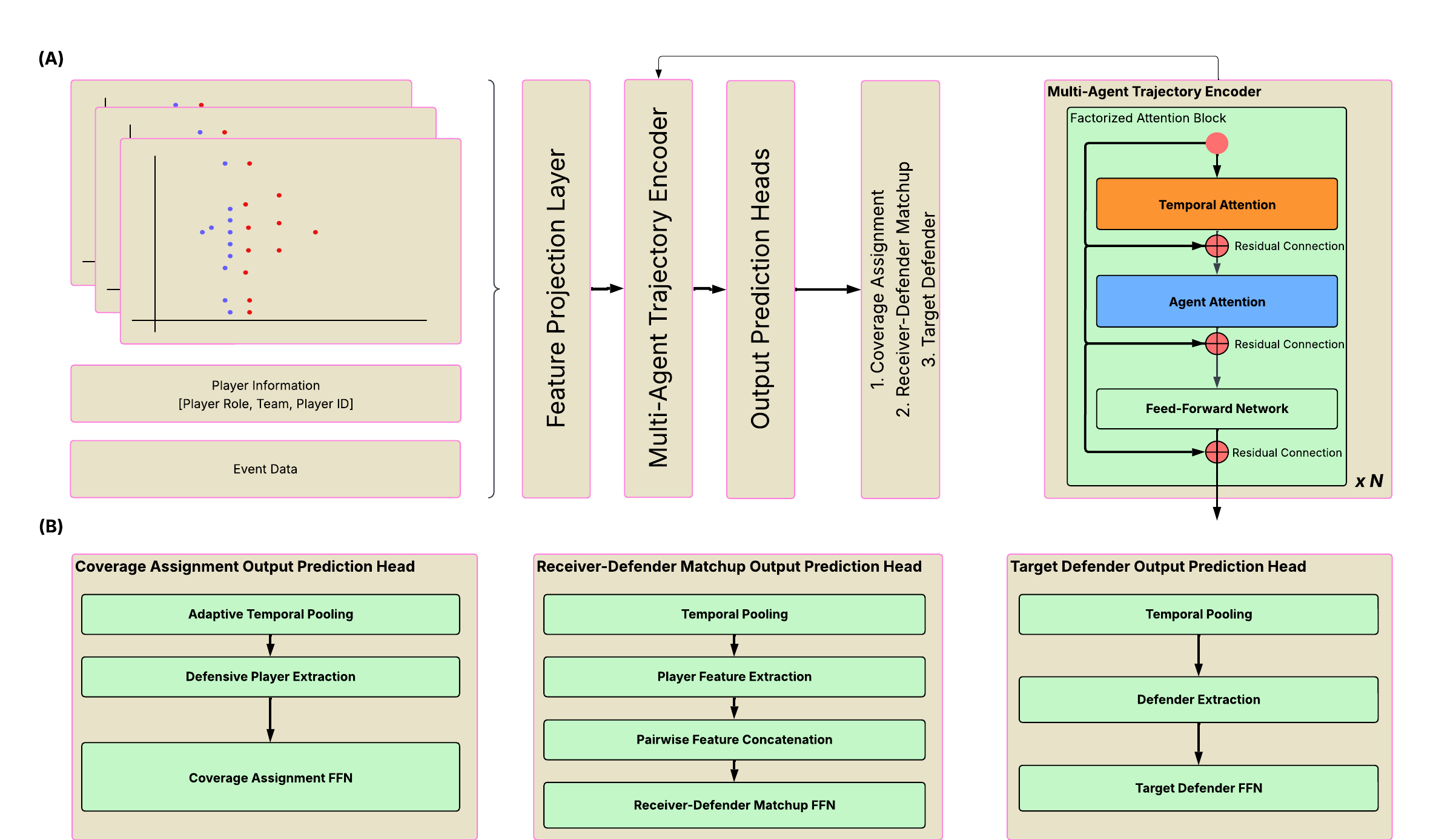}
\caption{Model architecture of the three coverage responsibility models. (A) General architecture that's shared amongst the 3 models (B) Different output heads for each model.}
\label{fig:model_arch}
\end{figure}

While each task shares this common model backbone, there are subtle nuances specific to each objective. The main differences are in the nature of the output classification which are highlighted in Figure~\ref{fig:model_arch}B. At a high level, all prediction heads apply a temporal pooling layer to get an average representation of each player at the play level. These player features are then extracted and fed to a feed-forward neural network (FFN) for final predictions.

For the coverage assignment task, only defensive player features are extracted and each feature set is fed to the coverage assignment FFN with an output shape of [num\_defenders, 20], where there are 20 different types of coverage assignments, including ``no assignments'' for pass rushers. For target defenders, only defensive player features are extracted, and the output of shape [num\_defenders + targeted\_receiver] contains probabilities for each defender being the primary target on a given play. For receiver-defender matchups, the features from eligible receivers and defenders are extracted, and we create a pairwise feature concatenation of all possible receiver-defender matchups. Then, the concatenated pairwise features are fed to a FFN and the output shape is [num\_defenders, num\_receivers + 1], corresponding to the probability that each defender is matched up to each receiver, including a class for ``no matchup''.

\subsection{Data Overview}

For all tasks, the NFL NGS data from seasons 2020 through 2024 were used, where seasons 2020-2023 were used for training and season 2024 data was used for testing. Broadly speaking, there are 5 data sources: tracking, play events, participation, Pro Football Focus (PFF) ground truth labels, and play data. The tracking data contains player and ball x,y coordinates within the field at a 10Hz sample rate. Events data contains game event markers such as snap, pass forward, and pass arrival, which are used to identify key events in a play. Participation data contains player roster information. PFF labels contain coverage responsibility ground truth annotation that includes individual coverage assignment, defender-receiver matchups, and target defender. The play data contains metadata such as play direction and field position.

\subsection{Data Processing}

The NGS dataset is processed into structured sequences suitable for transformer-based machine learning models by integrating data sources above and applying filtering to train the transformer model. First, we extract play sequences from -30 pre-snap frames at 10 Hz to pass forward frame, where frame 0 indicates the snap frame. After we extract the play sequence, we apply directional normalization, where all plays are normalized to run left-to-right regardless of the original play direction. In addition to tracking x and y coordinates, the NGS data also traces player orientation and movement direction. So, after normalizing overall play direction, we apply the same angular transformation on player orientation and movement direction vectors to ensure 0° represents rightward play movement. Then, we apply a series of filtering logic to remove low quality data. At the play level, we initially only include passing plays that contain valid snap and pass forward events, and we filter out plays that do not contain corresponding PFF annotations for each task.

\subsection{Training}

To enable temporal predictions throughout play development, we employed random trajectory truncation during training, using 11 fixed truncation strategies combined with random truncation as discussed in more detail in Appendix~\ref{sec:data_aug}. All three models employ cross-entropy loss as their fundamental training objective, with task-specific adaptations for the distinct prediction structures. Detailed loss function formulations and training configurations are provided in Appendix~\ref{sec:training_obj}.

\section{Results}

Four seasons from 2020 to 2023 were used for training, and season 2024 was used as the final test set. The ground truth labels for target defender, receiver-defender matchups, and coverage assignment were annotated at the play-level by human labelers, meaning each play received a single set of labels typically reflecting the state at the end of the play. This labeling approach, while practical for large-scale annotation, does not fully capture the dynamic nature of defensive coverage as it evolves throughout a play. For example, defenders may switch assignments, hand off receivers, or rotate into different zones as the play develops. However, because our model was trained on randomly truncated trajectories as described in Section 4.4 and Appendix A.1, the model learned to generalize and predict purely on trajectory and spatial patterns. The model is capable of generating predictions at a frame-by-frame level, producing coverage assignment, matchup, and target defender outputs at any point during the play's development. This temporal prediction capability allows us to evaluate model performance across different time points throughout the play, comparing predictions made with varying amounts of trajectory context—from pre-snap alignments through post-snap development to pass arrival. We leverage this capability in the following subsections by systematically evaluating each model across 11 distinct trajectory truncation strategies.

\subsection{Receiver-Defender Matchups}

\begin{table}
\caption{Receiver-defender matchup accuracy across different time frames.}
\label{tab:matchup}
\centering
\small
\begin{tabular}{lcccc}
\toprule
Time frame & Accuracy & F1 \\
\midrule
-30 to pass arrival & 0.894 & 0.894 \\
-20 to pass arrival & 0.894 & 0.894 \\
-10 to pass arrival & 0.894 & 0.894 \\
Snap to pass arrival & 0.893 & 0.893 \\
-20 to pass forward & 0.873 & 0.872 \\
-10 to pass forward & 0.873 & 0.872 \\
Snap to pass forward & 0.872 & 0.871 \\
-30 to pass forward & 0.872 & 0.871 \\
-20 to snap & 0.752 & 0.746 \\
-30 to snap & 0.751 & 0.746 \\
-10 to snap & 0.750 & 0.745 \\
\bottomrule
\end{tabular}
\end{table}

To evaluate temporal robustness and real-time deployment feasibility, we implemented an evaluation framework that assesses model performance across 11 distinct trajectory truncation strategies. Each strategy defines a temporal window by combining a start boundary with an end boundary. Start boundaries include 30, 20, or 10 frames before the snap, as well as the snap itself. End boundaries include the snap, pass forward, or pass arrival events. For example, one strategy uses frames from 30 frames pre-snap through pass arrival, while another uses only the snap frame through pass forward. This systematic approach allows us to assess how prediction accuracy varies with the amount of available trajectory context. Accuracy is computed per-defender, where a prediction is counted as correct if the receiver assigned to a given defender matches the ground truth label for that defender.

As shown in Table~\ref{tab:matchup}, the model achieves up to 89.4\% accuracy in predicting the correct receiver-defender matchups when using trajectory data through pass arrival. Importantly, this accuracy likely underestimates the model's true performance because the ground truth labels were assigned at the play level, reflecting a single snapshot of matchup assignments typically captured at the end of the play. On plays where matchups shift mid-play due to coverage rotations, route exchanges, or zone handoffs, the model may produce predictions that are correct for a given moment in the play but differ from the end-of-play label. In general, accuracy improves as more temporal context is provided: pre-snap trajectory data alone yields approximately 75\% accuracy, while including post-snap trajectories raises performance to 87\%+ across all configurations. This gap reflects the natural evolution of matchup assignments after the snap, and further reinforces that the model is effectively learning to track how coverage responsibilities develop in real time.

\subsection{Coverage Assignments}
The coverage assignment model employs the same 11 trajectory truncation strategies described in Section 4.1 to evaluate temporal robustness. The key differentiator in performance is whether the model has access to post-snap data. As shown in Table~\ref{tab:coverage}, all truncation strategies that include post-snap trajectories, ending at either pass forward or pass arrival, achieve approximately 92\% accuracy on coverage defenders. In contrast, strategies limited to pre-snap data yield approximately 88\% accuracy, a roughly 4-percentage-point gap. It is notable that the model can detect substantial signal from pre-snap alignments alone, and the performance gap between pre- and post-snap predictions may reflect deliberate deception by the defense, a topic we explore further in the Discussion section.

\begin{table}
\caption{Individual coverage assignment accuracy across different time frames.}
\label{tab:coverage}
\centering
\small
\begin{tabular}{lcccc}
\toprule
Time frame & Accuracy & F1 \\
\midrule
-30 to pass arrival & 0.918 & 0.879 \\
-20 to pass arrival & 0.918 & 0.879 \\
-10 to pass arrival & 0.918 & 0.878 \\
-10 to pass forward & 0.919 & 0.879 \\
-20 to pass forward & 0.918 & 0.878 \\
-30 to pass forward & 0.918 & 0.878 \\
Snap to pass forward & 0.917 & 0.877 \\
Snap to pass arrival & 0.917 & 0.877 \\
-30 to snap & 0.883 & 0.827 \\
-20 to snap & 0.883 & 0.827 \\
-10 to snap & 0.882 & 0.825 \\
\bottomrule
\end{tabular}
\end{table}
\subsection{Target Defender}
Unlike receiver-defender matchups and coverage assignments where there was no previous classification mechanism, the NFL NGS team was previously using a rule-based approach to determine target defenders. This metric leveraged the nearest defender to the targeted receiver at pass arrival. This metric provides a benchmark for comparison.

Another key difference is the application of rule-based logic that considers the situation surrounding the play, applied to augment predictions when probabilities are below 50\%. This post-processing step increases the accuracy of predictions by assigning a ``no target defender'' label when the defense is in a preventative alignment or when receptions occur behind the line of scrimmage. In other low-confidence cases, it defaults to the receiver-defender matchup model's assignment for targeted receiver classification. Target defender accuracy is calculated via the ratio between the number of plays with a correctly classified target defender over the total number of plays in the holdout season.

\begin{table}
\caption{Target defender classification accuracy against rule-based method and with and without post-processing rules.}
\label{tab:target}
\centering
\begin{tabular}{lc}
\toprule
Implementation & Accuracy \\
\midrule
Rules-Based Nearest Defender & 0.764 \\
Transformer & 0.862 \\
Transformer + Post-Processing Rules & 0.882 \\
\bottomrule
\end{tabular}
\end{table}

Overall, the model with post-processing rules achieves an 88.2\% accuracy in predicting the true target defender at the play-level. This deployed target defender classification process achieves about 12 percentage points of lift over the previous nearest defender approach.

\section{Discussion}

\subsection{Receiver-Defender Matchups}

The receiver-defender matchup model demonstrates robust performance across different temporal contexts, achieving up to 89.4\% accuracy when using trajectory data from pre-snap through pass arrival. Previously, receiver-defender matchups were approximated by simply determining who lined up against who at the snap~\cite{noauthor_nfl_nodate}. This ignored key dynamics of what happened after the snap, especially when the defense played zone coverage. In comparison to this previous distance-based algorithm, which was unable to capture the nuanced dynamics of coverage matchups throughout play development, the new model represents a significant improvement. Unlike simple proximity-based heuristics, the factorized attention-based transformer models utilize player trajectories and spatial information to best predict the corresponding matchups, which allows tracking of how matchup assignments evolve as players' routes develop.

In addition, a key strength of the transformer-based approach is its temporal robustness. The model maintains approximately 75\% accuracy using only pre-snap data (from -30, -20, or -10 frames to snap), which provides valuable insights into defensive pre-snap alignment and initial coverage intentions. Performance improves substantially to 87-89\% when post-snap trajectory data is included, reflecting the model's ability to confirm matchup assignments as the play unfolds.

\begin{figure}
\centering
\includegraphics[width=\textwidth]{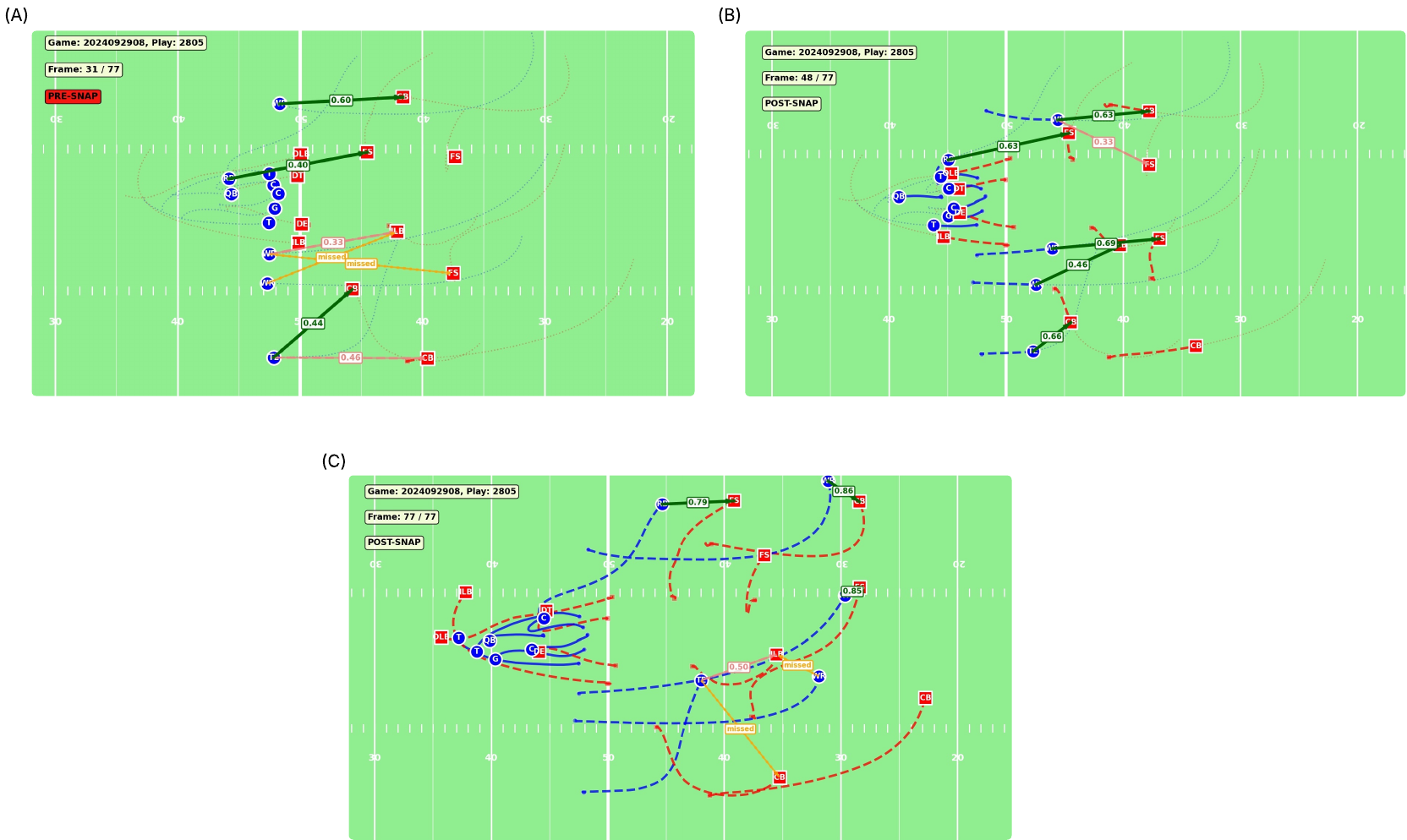}
\caption{Receiver-defender matchup results over time. Receiver-defender matchup predictions at (A) pre-snap, (B) mid-play, and (C) pass arrival.}
\label{fig:matchup_evolution}
\end{figure}

Figure~\ref{fig:matchup_evolution} illustrates how matchup predictions evolve throughout the play. At pre-snap (Figure~\ref{fig:matchup_evolution}A) and at full play (Figure~\ref{fig:matchup_evolution}C), the model incorrectly predicts 2 out of 5 matchup assignments. However, at mid-play (Figure~\ref{fig:matchup_evolution}B), the model achieves perfect accuracy, correctly identifying all 5 matchups. This temporal variation is particularly evident in the TE-CB matchup shown at the bottom of Figures~\ref{fig:matchup_evolution}A and~\ref{fig:matchup_evolution}B. Initially, the cornerback appears to be covering the tight end, but as the play develops, the tight end cuts inside while the cornerback transitions to zone coverage. The ground truth annotations were likely labeled based on mid-play context, which explains why the model's full-play predictions appear reasonable despite conflicting with the ground truth label.

This discrepancy highlights a key advantage of the model: it adapts its predictions dynamically based on available context, whereas human annotators were constrained to provide a single label per play. The model's capacity for generating temporal predictions throughout play development reveals valuable tactical insights, such as identifying when defenders execute coverage handoffs mid-play. Training the model on randomly truncated plays enabled it to learn robust spatio-temporal pattern recognition across varying trajectory lengths, from brief pre-snap sequences to complete play development, and this allows the model to flexibly make predictions with any amount of play context.

These matchup predictions serve as a critical foundation for downstream metrics and analyses, including double coverage identification and tight coverage evaluation. By reliably attributing which defenders are responsible for covering specific receivers, we enable more sophisticated defensive performance evaluation and tactical insights.

\subsection{Coverage Assignment}

For the coverage assignment task, the model generates predictions for all 11 defenders on every play. On approximately 70\% of 2024 plays, all coverage defenders were correctly classified at pass arrival. An additional 20\% had between 1 and 3 incorrect predictions---``judgment calls'' representing annotation ambiguity, quick passes where coverage cannot fully develop, or genuinely borderline situations.

\begin{figure}
\centering
\includegraphics[width=\textwidth]{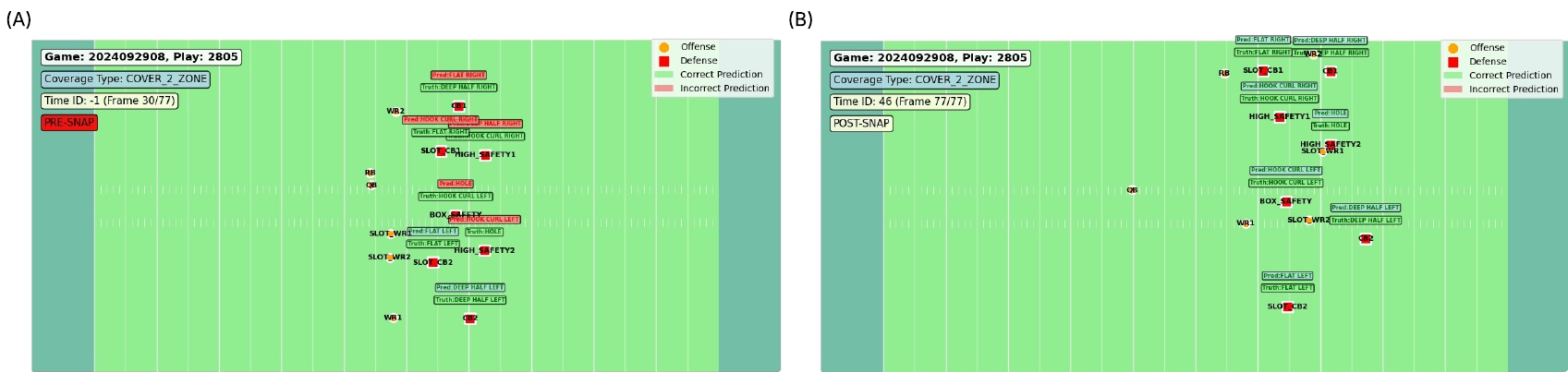}
\caption{Coverage assignment prediction over time. From left to right, the frames show snapshots of predictions (A) 3 seconds before the snap and at (B) pass arrival.}
\label{fig:coverage_evolution}
\end{figure}

Figure~\ref{fig:coverage_evolution} shows an example where all defenders' assignments were correctly classified at pass arrival. This particular play is illustrative of the model's flexibility because in conventional Cover 2, the two high safeties would take the deep left and right halves, outside cornerbacks cover the flats, and slot corners cover hook/curl routes in the middle of the field. The pre-snap predictions (Figure~\ref{fig:coverage_evolution}A) show that the model has internalized this concept. However, as the play unfolds, the defensive backs rotate such that the outside cornerbacks drop back to occupy the deep left and right halves, the high safeties step forward into the middle of the field, and the slot corners shift to the flats. By the time of pass arrival (Figure~\ref{fig:coverage_evolution}B), the model has detected this disguise and updated its predictions accordingly. This example also presents an added benefit of classifying individual coverage responsibilities, as it captures the ``Invert Cover 2'' team coverage scheme. These adjustments and more specific coverage concepts are unlocked, moving into more detailed classifications of team coverage schemes beyond the general families.

\begin{figure}
\centering
\includegraphics[width=\textwidth]{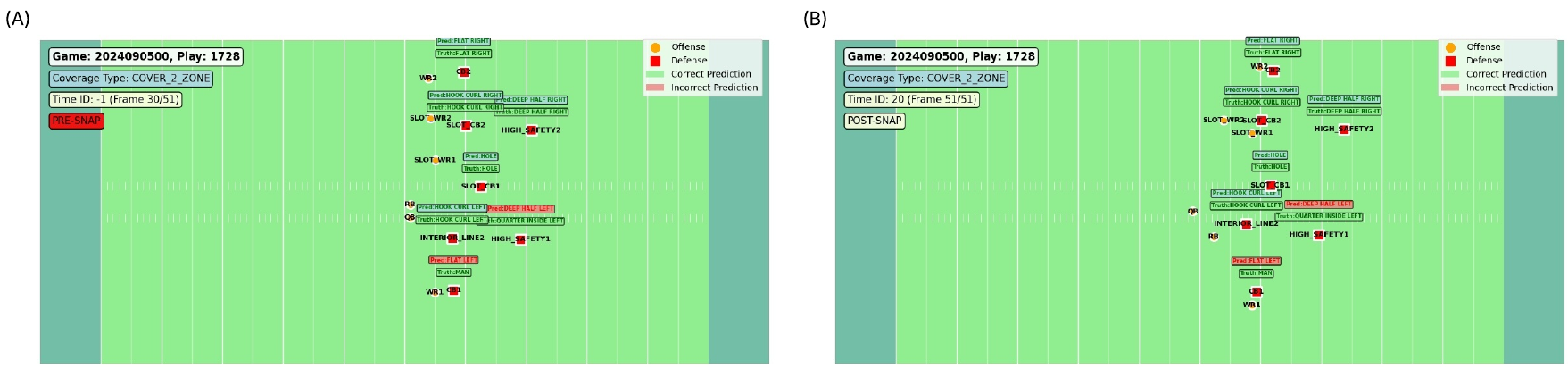}
\caption{Misclassified individual assignments due to potential mislabeled samples (HIGH\_SAFETY1) and/or genuine ambiguity in man vs. zone assignments (CB1). (A) Coverage assignment predictions at pre-snap and (B) coverage assignments at full play both show misclassified predictions. This example shows the model's bias toward symmetry and respecting the overall coverage scheme.}
\label{fig:coverage_misclass}
\end{figure}

The play shown in Figure~\ref{fig:coverage_misclass} contains 2 incorrect classifications. At just 2.1 seconds from snap to pass arrival, the compressed timeline adds inherent difficulty, yet the model still correctly identifies the overall Cover 2 shell. One of the misclassifications involves the deep safety on the left side: the model predicts a left deep half zone assignment, while the human label indicates quarters inside left. This discrepancy is likely attributable to human labeling error, as the quarters inside left label violates the charting guidelines since no teammate is assigned the outside quarters area on that side of the field. In reality, the available label set may be insufficient to describe what this safety is doing; upon review of the all-22 film, the safety appears to be ``poaching'' any route from the three-receiver side that crosses the field, while the cornerback to his side locks onto the isolated receiver in man coverage. The model has learned that most coverage schemes exhibit spatial symmetry, such as flat left and flat right assignments on opposite sides of the field, and defaults to this principle unless the play dynamics indicate otherwise.

The next tier of misclassifications generally stems from scheme-level disagreement: because individual assignments are tightly coupled to the overall shell, a single misalignment cascades into multiple per-defender errors with an outsized effect on accuracy. These scheme disagreements often occupy a gray area where the model's predictions are plausible, if not preferable, to the human-generated labels. The most extreme cases (~3\% of plays, with 6+ errors) arise from disagreements on specialized schemes such as Prevent, where the model assigns Prevent labels to all coverage defenders, concentrating misclassifications on a single play.

\begin{figure}
\centering
\includegraphics[width=\textwidth]{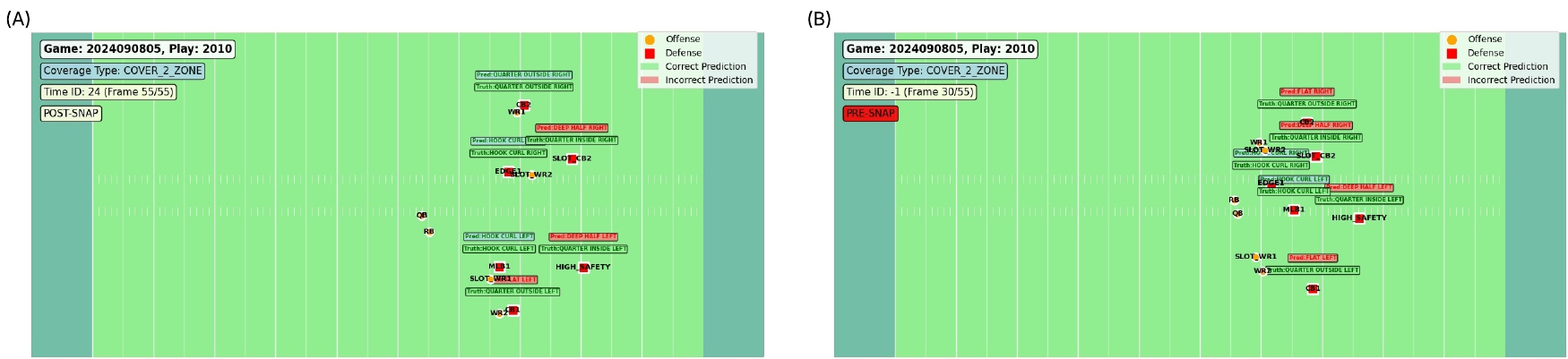}
\caption{Misclassified individual assignments due to disagreement on overall coverage scheme, leading to a cascade of predictions that do not align with the human labels, but are internally consistent within the context of the predicted Cover 2 shell.}
\label{fig:coverage_scheme}
\end{figure}

Figure~\ref{fig:coverage_scheme} provides a representative example. The play contains 4 ``incorrect'' individual assignments at pass arrival, all stemming from ambiguity in the overall coverage scheme. The human labelers classified the play as Cover 4, while the model identified it as Cover 2. This uncertainty is compounded by a quick-game passing concept in which the quarterback releases the ball within 2 seconds of the snap, leaving little time for the coverage to fully develop. The human labelers may have inferred deep quarters coverage based on the outside corners' initial alignment depth. In contrast, the model observes that the two deep safeties gain significantly more depth than the corners, who remain relatively stationary, suggesting flat responsibilities consistent with a Cover 2 scheme. Critically, the model's individual assignments are internally consistent with its identified scheme, demonstrating that it considers the relationships between defenders within the broader coverage context rather than classifying each defender in isolation.

Overall, the model consistently produces plausible predictions. In cases of disagreement with the ground truth labels, our inspection suggests that the discrepancy reflects either a genuine gray area in the labeling or an error in the human-generated annotations. This interpretation is supported both by anecdotal review of ``incorrect'' classifications and by the model's Top-2 accuracy of 97\% across 20 possible assignment classes. In other words, the reported 89\%+ accuracy may still understate the model's true performance, as many of the ``errors'' represent defensible alternative interpretations rather than genuine prediction failures.

\subsection{Target Defender}

\begin{figure}
\centering
\includegraphics[width=\textwidth]{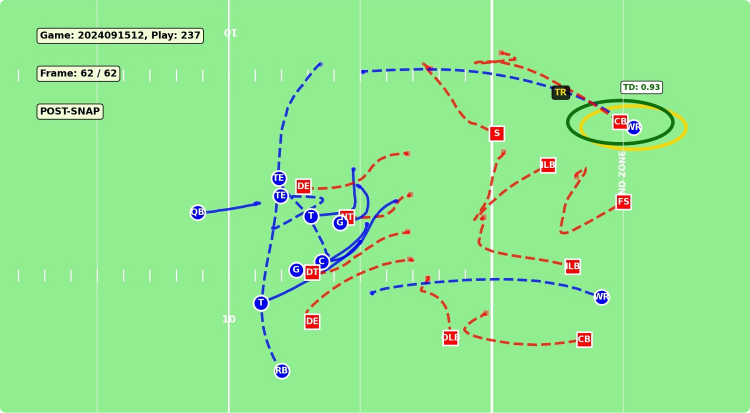}
\caption{Play level target defender prediction illustrated at pass forward.}
\label{fig:target_defender}
\end{figure}

The factorized attention-based transformer model demonstrates significant improvement over the previous rules-based approach for identifying the target defender, achieving 88.2\% accuracy compared to 76.4\% for the nearest defender heuristic. This 12-percentage-point lift represents a substantial advancement in the ability to identify which defender the quarterback is targeting across various coverage schemes, and it is driven by a key property of the factorized attention mechanism.

Most notably, the transformer architecture incorporates temporal context that the rules-based approach cannot. While the nearest defender heuristic only considers defender proximity at a single point in time, the model analyzes the full sequence of player movements leading up to the pass. As illustrated in Figure~\ref{fig:target_defender}, the model correctly predicts the targeted defender by integrating both trajectory dynamics and final positioning. More broadly, the model learns to detect subtle cues in defender movement patterns and receiver route concepts that signal which defender is the true target of the play.

The strong performance of the target defender model has important implications for defensive player evaluation. Accurately attributing coverage responsibility to the correct defender, rather than approximating credit based on physical proximity, is essential for fair and meaningful assessment of individual performance in coverage. This capability lays the groundwork for more reliable player grading and schematic evaluation across the league.

\subsection{Disguise Rate}

Predicting pre-snap coverage assignments in the NFL offers valuable insights into one of football's most critical strategic elements: defensive deception~\cite{noauthor_exposing_nodate,noauthor_nfl_nodate-1}. By accurately identifying coverage schemes before the snap, analysts can quantify how effectively defenses disguise their true intentions, revealing the gap between what they show and what they execute. This capability provides offensive coordinators and quarterbacks with a significant tactical advantage, enabling them to better anticipate defensive rotations and make better informed pre-snap reads.

\begin{figure}
\centering
\includegraphics[width=\textwidth]{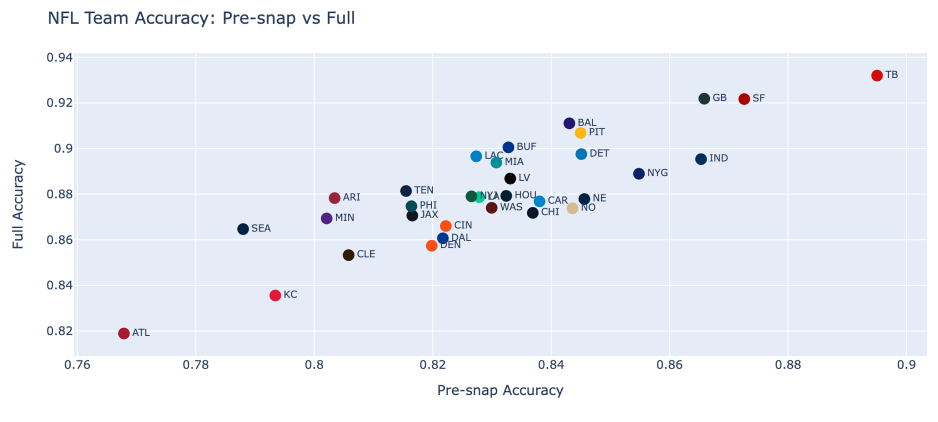}
\caption{Disguise rate was approximated by calculating pre-snap coverage assignment accuracy and full accuracy.}
\label{fig:disguise}
\end{figure}

Because the coverage assignment model was trained to see a range of truncated plays, including combinations of pre- and post-snap trajectories, it can make robust predictions before the ball is snapped. The model could make an ``accurate'' prediction at pre-snap based on the defensive players' spatial arrangements, but the players may move into a completely different formation as soon as the ball is snapped. Accuracy at pre-snap predictions can provide a proxy for deception, as defenses attempt to disguise the coverage as the quarterback makes his initial reads. The teams that have the lowest pre-snap accuracy can be thought of as those demonstrating the most deceptive pre-snap. One notable example is the Kansas City Chiefs as shown in Figure~\ref{fig:disguise}; Defensive Coordinator Steve Spagnuolo is famous for his bold stunts and exotic blitz packages. At the other end of the spectrum, Tampa Bay and San Francisco appear to show more predictable schemes. While the Tampa Bay Buccaneers Head Coach Todd Bowles is also known for his more unpredictable blitz packages, the high accuracy rate here might indicate that his coverage rotations are easily identified by the model based on his secondary's timing of their movement before the snap.

This concept of ``disguise rate'' and the difference between pre- and post-snap labels can unlock other advanced capabilities, like instances of inverted coverages.

\subsection{Double Coverage Rate}

Another example of a derivative metric that can be calculated with the three models presented here is double coverage rate. An instance of a double coverage is defined as a play where at least 2 defenders playing man coverage have a primary matchup against the same receiver. We can therefore derive these cases by combining the outputs from the Receiver-Defender Matchups and Coverage Assignment models, where we define a double coverage as greater than or equal to 2 defenders with man coverage assignments matched to a single receiver.

True double coverage is a relatively rare phenomenon reserved for specific scenarios and for dominant receivers. For example, when analyzing the human-labeled ground truth, there were only 1,179 instances of double coverage across 21,567 eligible pass plays during the 2024 season (5.5\%). The objective of this analysis is to calibrate the sensitivity of the upstream models to ensure that the global population of model-defined instances of double coverage approximately matches the ground truth. We then aggregate the results at the team defense and receiver level to find Double Coverage Rates across the league.

\begin{figure}[h!]
\centering
\includegraphics[width=\textwidth]{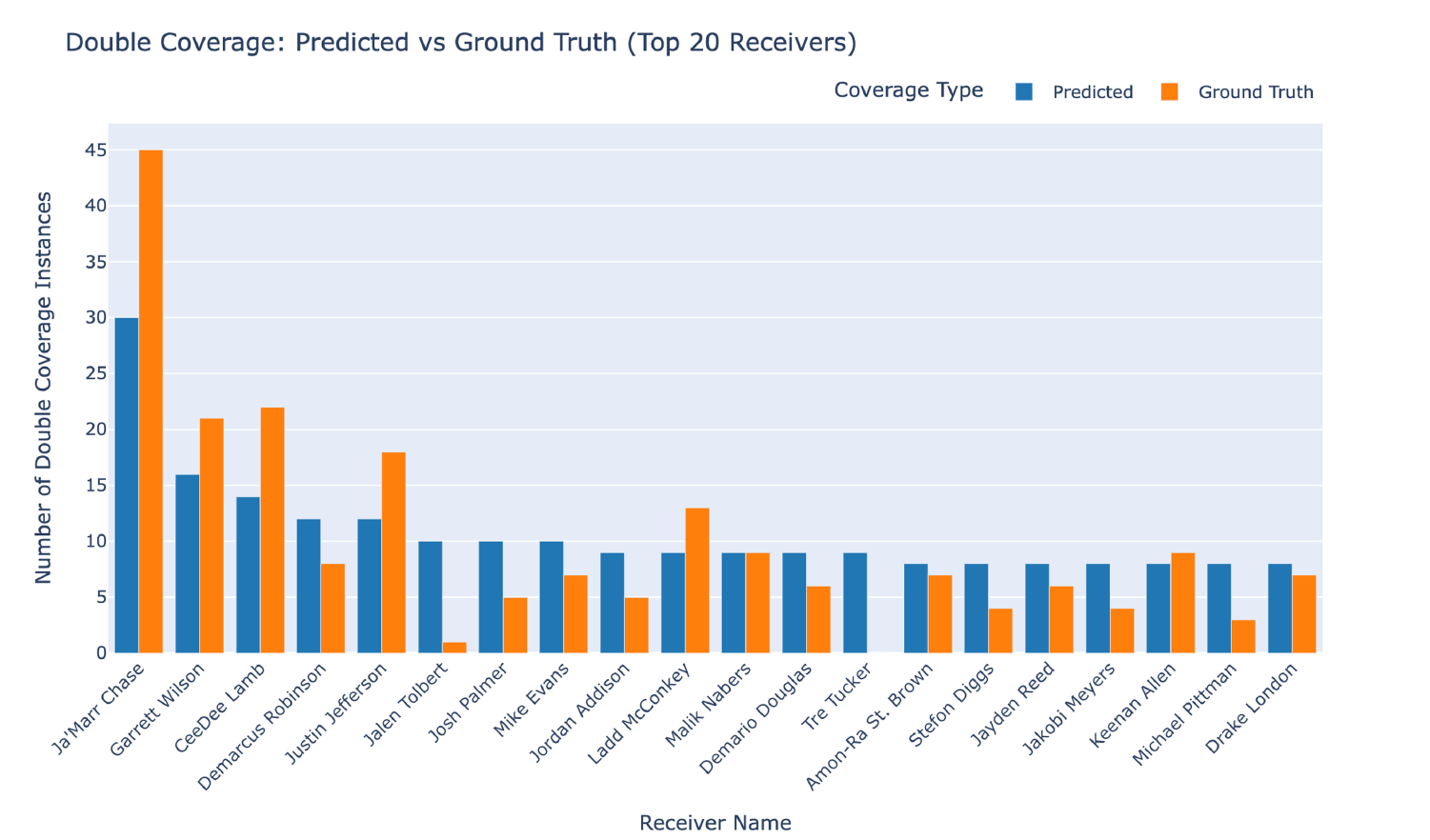}
\caption{Comparison of Double coverage rate per player over the 2024 season between the prediction and ground truth.}
\label{fig:double_coverage}
\end{figure}

Using the coverage assignment and defender-receiver matchup models, the number of double coverages was calculated per receiver. The results are shown in Figure~\ref{fig:double_coverage}. The top-20 most double-covered receivers according to the model show general alignment with the human labels and generally pass the ``eye test'' in terms of the names at the top of the list. It comes as no surprise that Ja'Marr Chase, Garrett Wilson, CeeDee Lamb, and Justin Jefferson are in the top 5.

Outside of the top 5 double covered receivers, there were a few notable surprises like Jalen Tolbert and Tre Tucker. For Tre Tucker specifically, we found 9 instances of double coverage whereas the human labels identified 0. It is plausible that there is a ``reputation bias'' in the human labels compared to the model, where labelers are more likely to identify double coverage on well-known, dominant players like Chase, Lamb, and Jefferson. This bias may truly be more in line with what NFL defensive coordinators are game-planning for, but on the other hand, the model may be more objective in capturing the on-field realities or ``leading indicators'' about underrated receivers. It's also important to caveat this analysis with the reality that this is a rare phenomenon, and disagreements on a handful of ``judgment call'' plays can look drastic with the small sample size. For example, even Mike Evans saw only 10 double coverages over the whole season.

\section{Conclusion}
In this paper, we demonstrated the application and effectiveness of factorized attention-based transformers for classifying individual coverage responsibilities across three tasks: target defender identification, receiver-defender matchup prediction, and coverage assignment classification. While the models achieve approximately 89\%+ accuracy, true accuracy may be higher given human labeling ambiguity, with the model's predictions consistently defensible and often preferable to human annotations in gray-area situations. A key contribution is the model's ability to generate frame-by-frame predictions, enabling analysts to observe how defensive responsibilities evolve throughout the play rather than relying on static end-of-play labels. These models together also enable derivative metrics such as disguise rate and double coverage rate. The models are deployed for NFL broadcast productions, powering real-time coverage graphics during live games, and the architecture could be extended to other dynamic team sports such as basketball and soccer. Future work includes systematic ablation studies isolating the contributions of agent attention, temporal attention, truncation augmentation, and the upstream coverage feature, as well as comparisons against sequence modeling baselines such as GRU and TCN architectures.

\bibliographystyle{splncs04}
\bibliography{coverage_resp}

\appendix

\section{Appendix}

\subsection{Data Augmentation}
\label{sec:data_aug}

To extend the transformer model's capabilities to predict better at different times on any given play, a data augmentation with random truncation scheme was applied to help the model learn more effectively. For any given play, there are 11 fixed truncation strategies where the play can be truncated to any of the following combinations where the first item in the tuple represents the start frame, and the second item in the tuple represents the end frame. The following combinations were used: (-30, snap), (-30, pass forward), (-30, pass arrival), (-20, snap), (-20, pass forward), (-20, pass arrival), (-10, snap), (-10, pass forward), (-10, pass arrival), (0, pass forward), and (0, pass arrival), where 0 represents the snap frame and negative frame numbers denote frames prior to snap. In addition to the 11 fixed truncation strategies, a truly random truncation strategy was also added, which will truncate the start and end frames anywhere from -30 to pass arrival. With the two fixed and random truncation strategies in place during training, each play is truncated with either fixed truncation or random truncation at a 60\% and 40\% split, respectively.

\subsection{Training Objectives and Loss Functions}
\label{sec:training_obj}

All three models employ cross-entropy loss as their fundamental training objective due to their shared nature as classification tasks. Each model processes multi-agent trajectory data through factorized transformer encoders that separate temporal and spatial attention mechanisms, ultimately producing logits that are compared against ground truth labels using cross-entropy loss.

While the three models share the same fundamental loss function, the three models differ in how cross-entropy is applied due to their distinct prediction structures and task requirements. The target defender model performs single-label classification across all defenders on a play, computing cross-entropy over a categorical distribution where exactly one defender is selected as the primary target. The receiver-defender matchup model, in contrast, computes cross-entropy independently for each defender, treating each as a separate multi-class classification problem over receivers plus a ``no matchup'' class. The coverage assignment model extends this further by performing independent multi-class classification for each of the 11 defensive players, predicting 20 coverage responsibilities (e.g. hook curl left, deep right, etc.) with cross-entropy computed per defender and averaged across the defensive unit.

\subsubsection{Receiver-Defender Matchups}

For each defender, the model predicts which eligible receiver they are matched to in a prediction format of (batch\_size, num\_defenders, num\_receivers + 1), where the additional dimension represents a ``no matchup'' class for defenders not assigned to cover any receiver. For defenders with no coverage assignments, indicated by zero-sum rows in the transposed target matrix, the ``no matchup'' class label is set to 1.0, while defenders with valid assignments retain their original one-hot encoded receiver assignments. Then, the loss for matchup is computed as:
\begin{equation}
\mathcal{L}_{\text{matchup}} = -\frac{1}{D}\sum_{d=1}^{D}\sum_{r=0}^{R} z_{d,r}^* \log\left(\frac{\exp(Z_{d,r})}{\sum_{r'=0}^{R}\exp(Z_{d,r'})}\right)
\label{eq:matchup}
\end{equation}
where $Z_{d,r}$ is the logit score for defender $d$ covering receiver $r$, $R$ is the number of eligible receivers, and $z_{d,r}^*$ is the ground truth receiver assignment for defender $d$.

\subsubsection{Coverage Assignment}

For individual coverage assignments, the training objective is to assign a single assignment for each of the 11 defenders on a given play. The prediction output shape is (batch\_size, num\_defenders, 20), where there are 19 total coverage assignment classes with 1 no assignment class. The goal of training is to minimize cross entropy loss for the 20 possible coverage assignments, defined as:
\begin{equation}
\mathcal{L}_{\text{coverage}} = -\frac{1}{D}\sum_{d=1}^{D}\sum_{c=1}^{C} \mathbb{1}[c = c^*] \log\left(\frac{\exp(z_{d,c})}{\sum_{c'=1}^{C}\exp(z_{d,c'})}\right)
\label{eq:coverage}
\end{equation}
where $z_{d,c}$ is the logit score for each defender having coverage assignment $c$, $C$ is the number of coverage classes, and $c^*$ is the ground truth coverage class for defender $d$.

\subsubsection{Target Defender}

For passing plays, the quarterback attempts to beat a defender to make a successful completion to an eligible receiver and advance offensive field position. The target defender model determines which of the 11 defenders present in each passing play is the primary defender that quarterback is looking to beat. For each play, only one defender is the true target defender.

The model outputs logits for each player with the following shape: (batch\_size, num\_defenders + targeted receiver). These are then masked to only consider defensive players as valid candidates by setting the offensive player's (targeted receiver) logit to a large negative value (-1e9) before the softmax operation. The training process minimizes the following cross-entropy loss function where $N$ is the number of players, $z_j$ is the logit for player $j$, and $z_d$ refers to the logit of the ground truth primary defender:
\begin{equation}
\mathcal{L}_{\text{target}} = -\log\left(\frac{\exp(z_d)}{\sum_{j=1}^{N}\exp(z_j)}\right)
\label{eq:target}
\end{equation}

\subsection{Training Hyperparameters}
Hyperparameters used for training the final models are shown in Table ~\ref{tab:hyperparams}.

\begin{table}
\caption{Training Optimization and Hyperparameter settings}
\label{tab:hyperparams}
\centering
\small
\begin{tabular}{p{3cm}p{3.5cm}p{3.5cm}p{3.5cm}}
\toprule
\textbf{Training Task} & \textbf{Target Defender} & \textbf{Receiver-Defender Matchups} & \textbf{Coverage \newline Assignments} \\
\midrule
Batch Size & 16 & 16 & 16 \\
\midrule
Num Epochs & 100 & 200 & 150 \\
\midrule
Learning Rate and \newline Scheduler & OneCycleLR ($max\_lr = 2e^{-4}$, 
\newline $total\_steps = 
\newline train\_len * 100$, 
\newline $pct\_start=0.1, 
\newline div\_factor=10.0$, 
\newline $final\_div\_factor=
\newline 100.0$)
& OneCycleLR ($max\_lr=2e^{-4}$, 
\newline $total\_steps = 
\newline train\_len * 200$, 
\newline $pct\_start=0.1, 
\newline div\_factor=10.0$, 
\newline $final\_div\_factor = 
\newline 100.0$) 
& Cosine Annealing Warm Restarts 
\newline $init\_lr = 2e^{-5}$
\newline $T_0 = 15$
\newline $T_mult = 1$
\newline $eta\_min=2e^{-7}$ \\
\midrule
Optimizer & AdamW \newline weight decay=1e-4, \newline betas=(0.9, 0.999) & AdamW \newline weight decay=1e-4, \newline betas=(0.9, 0.999) & AdamW \newline weight decay=1e-4, \newline betas=(0.9, 0.999) \\
\midrule
Num Heads & 4 & 4 & 8 \\
\midrule
Num Layers & 3 & 3 & 6 \\
\midrule
Dropout & 0.1 & 0.2 & 0.1 \\
\bottomrule
\end{tabular}
\end{table}

\end{document}